\documentclass[11pt]{article}
\usepackage{acl}
\usepackage{times}
\usepackage{latexsym}
\usepackage[T1]{fontenc}
\usepackage[utf8]{inputenc}
\usepackage{microtype}
\usepackage{inconsolata}
\usepackage{graphicx}
\usepackage{amsmath}
\usepackage{booktabs}
\usepackage{makecell}
\usepackage{enumitem}
\usepackage{flafter}
\usepackage[table]{xcolor}
\usepackage[most]{tcolorbox}

\definecolor{pframe}{HTML}{3C4A5C}
\definecolor{pback}{HTML}{F7F8FA}
\definecolor{plabel}{HTML}{3C4A5C}
\definecolor{pslot}{HTML}{1F5FA8}

\newcommand{\ph}[1]{\textcolor{pslot}{\texttt{\{#1\}}}}
\newcommand{\psec}[1]{\par\medskip{\footnotesize\sffamily\bfseries\color{plabel}#1}\par\smallskip}

\newtcolorbox{pslotbox}[1]{enhanced, colback=white, colframe=gray!30,
  boxrule=0.4pt, arc=2pt, left=6pt, right=6pt, top=4pt, bottom=4pt,
  fontupper=\small, title={#1}, fonttitle=\footnotesize\sffamily\bfseries,
  coltitle=plabel, colbacktitle=gray!10, toptitle=1pt, bottomtitle=1pt}

\definecolor{headergray}{HTML}{DCE6F2}   
\definecolor{rowgray}{HTML}{F5F8FC}      
\definecolor{overallgray}{HTML}{E8EEF7}  

\title{Scalaris: A Scalable Enterprise Memory Architecture for Conversational AI Assistants}

\author{
  Devarsh Gandhi \enspace Sarthak Pawar \enspace Hemprasad Badgujar \enspace Vaibhav Rathore \\  \enspace
  \textbf{Siddharth Sureka} \enspace \textbf{Neha Bhargava} \\
  Motilal Oswal Financial Services Ltd \\
  \texttt{\{devarsh.gandhi, sarthak.pawar, hemprasad.badgujar,} \\
  \texttt{rathore.vaibhav, siddharth.s, neha.bhargava\}@motilaloswal.com}
}

\begin{document}
\maketitle

\begin{abstract}
Enterprise conversational systems require memory to personalize interactions, retain information across sessions, and maintain conversational coherence. At enterprise scale, however, memory must not only retrieve relevant information accurately but also provide it efficiently to the answer-generation model. Existing memory systems face a trade-off between retrieval accuracy and context efficiency. Retrieving more context can improve recall but increases inference cost and latency. We present Scalaris, a scalable enterprise memory architecture that addresses this trade-off through a unified long-term and short-term memory design. We propose that user interactions and assistant responses possess distinct characteristics and roles in a conversation: user interactions capture evolving knowledge about the user and provide value across sessions, whereas assistant responses must be stored efficiently so that only information relevant to the current query is retrieved. We therefore treat them separately and support each with specialized representations and ingestion/retrieval pipelines. Scalaris reaches 92.2\% accuracy on LongMemEval-S and 90.77\% on LoCoMo using Qwen3-32B, state-of-the-art results on both benchmarks, while supplying the answer-generation model with approximately 10$\times$ fewer retrieved context tokens than Hindsight, a leading memory system for conversational AI. When integrated with our proprietary closed-source AI assistants, Scalaris improved overall response quality by 36\%. Together, these results establish Scalaris as a cost-effective framework for building scalable, personalized conversational AI systems.
\end{abstract}

\section{Introduction}

The emergence of large language models has enabled organizations to deploy conversational AI systems across user-facing applications such as customer support and sales assistance~\citep{yi2024surveyllmdialogue}. Memory is essential to these systems for maintaining continuity across conversations and personalizing interactions~\citep{wu2025humanmemory}. However, simply providing an assistant with prior message history becomes limiting as conversations grow longer and span multiple sessions, since model performance degrades over long contexts~\citep{liu2024lost}. Existing memory frameworks have therefore explored mechanisms for selectively storing and retrieving information from past conversations~\citep{zhang2024memorysurvey}, but they often face a trade-off between retrieval accuracy and context efficiency. Zep~\citep{rasmussen2025zeptemporalknowledgegraph}, for example, retrieves relatively little context but achieves lower accuracy on long-term memory benchmarks, whereas Hindsight~\citep{latimer-etal-2026-hindsight} achieves substantially higher accuracy while retrieving considerably more. An effective enterprise memory system must therefore retrieve the information required to answer a query accurately without unnecessarily increasing the context supplied to the answer-generation model~\citep{lewis2020retrieval, shi2023distracted, chhikara2025mem0buildingproductionreadyai}.

\begin{figure}[!t]
    \centering
    \includegraphics[width=\columnwidth]{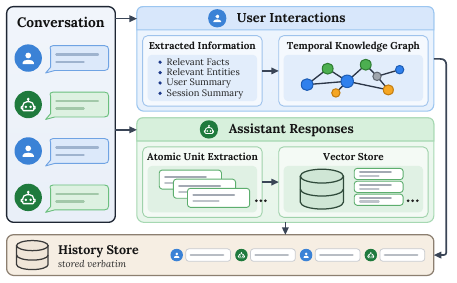}
    \caption{Overview of Scalaris's source-aware memory organization. User interactions are processed to maintain relevant facts, relevant entities, and user and session summaries in a temporal knowledge graph. Assistant responses are decomposed into independently retrievable atomic units and stored in a vector store. In parallel, both user interactions and assistant responses are preserved verbatim in a shared history store.}
    \label{fig:teaser}
    \vspace{-1.5em}
\end{figure}

Conversational memory must also operate across two different timescales. Information such as user preferences, interests, and circumstances may remain useful across sessions and must evolve as new interactions occur~\citep{xu2025amemagenticmemoryllm, zhang2024personalization}, creating the need for long-term memory. At the same time, an assistant must retain the immediate conversational state required to interpret follow-up questions, resolve references, and maintain coherence within an ongoing session~\citep{packer2024memgptllmsoperatingsystems, yu-etal-2026-agentic}, motivating short-term memory. A scalable conversational memory architecture must therefore support both persistent cross-session knowledge and immediate within-session context.

Existing systems typically place all conversational turns under a shared memory representation despite the different roles of user interactions and assistant responses. User interactions contain persistent facts, preferences, and evolving relationships that remain valuable across sessions; assistant responses are largely contextual and often obsolete outside their original dialogue. User interactions therefore benefit from relational representations that capture evolving user knowledge, while assistant responses are better stored as independently retrievable atomic units, so that only the relevant portion of a response is recalled rather than the whole.

\noindent \textbf{Our Proposal:} We present Scalaris, a unified long-term and short-term memory architecture designed around this distinction. As illustrated in Figure~\ref{fig:teaser}, Scalaris maintains separate memory representations for user interactions and assistant responses. User interactions are organized in a temporal knowledge graph that maintains relevant facts, entities, and evolving user information, together with user and session summaries. Assistant responses are decomposed into independently retrievable atomic units and stored in a vector database, allowing only information relevant to the current query to be retrieved. In parallel, user interactions and assistant responses are preserved verbatim in a shared history store for recent conversational context. By providing specialized ingestion and retrieval pipelines for these different forms of memory, Scalaris aims to preserve the information needed for accurate recall while limiting unnecessary retrieved context.

Our main contributions are as follows:

(1) We present Scalaris, a unified long-term and short-term memory framework for building scalable, enterprise-ready conversational AI assistants.

(2) We propose that user interactions and assistant responses exhibit distinct characteristics that call for different memory representations, and we support each with a specialized ingestion and retrieval pipeline.

(3) We achieve state-of-the-art performance on both LongMemEval-S (92.2\%) and LoCoMo (90.77\%) using the relatively small Qwen3-32B model, while requiring approximately 10$\times$ fewer retrieved context tokens than Hindsight.

\section{Related Work}
\label{sec:related}

\noindent \textbf{Long-Term and Knowledge-Graph Memory Representations.}
Long-term memory frameworks let conversational agents retain information across sessions through varied representations. Knowledge-graph approaches such as Zep~\citep{rasmussen2025zeptemporalknowledgegraph} and Supermemory~\citep{supermemory} capture entities and their evolving relationships over time, while Mem0~\citep{chhikara2025mem0buildingproductionreadyai} combines graph components with selective consolidation. LangMem~\cite{langmem} provides a general-purpose toolkit for extracting and consolidating memories. Unit-based and reflective models such as A-Mem~\citep{xu2025amemagenticmemoryllm} treat memory as interconnected units that update as new interactions occur; RGMem~\citep{tian2026rgmemrenormalizationgroupinspiredmemory} derives higher-level user profiles from episodic logs, and RMM~\citep{tan-etal-2025-prospect} applies reflective management to long-term dialogue. Others target efficiency through different mechanisms: SimpleMem~\citep{liu2026simplememefficientlifelongmemory} compresses conversations into compact semantic units, FraCom~\citep{ke-etal-2025-flexibly} composes fine-grained fragments at retrieval time, RecMem~\citep{dai2026recmemrecurrencebasedmemoryconsolidation} optimizes memory construction cost, and Zero-Mem~\citep{xiao2026zeromem} eliminates LLM calls during memory operations through deterministic memory management.

\noindent \textbf{Temporal Organization and Hierarchical Architectures.}
Operating across immediate and multi-session contexts requires tracking temporal evolution and balancing short- against long-term stores. MemGPT~\citep{packer2024memgptllmsoperatingsystems} introduced hierarchical, OS-like memory management, while AgeMem~\citep{yu-etal-2026-agentic} learns actions across long- and short-term tiers. To handle temporal shifts over long interaction spans, TiMem~\citep{li-etal-2026-timem}, APEX-MEM~\citep{banerjee-etal-2026-apex}, and THEANINE~\citep{ong-etal-2025-towards} structure information around timelines and state updates. Complementary work explores content-based categorization: Hindsight~\citep{latimer-etal-2026-hindsight} organizes memory into world knowledge, agent experience, entity observations, and beliefs. This implicitly separates assistant-originated opinions from user observations, but the division follows from epistemic role rather than a deliberate partition by the origin of the information.

\noindent \textbf{Partitioning User Interactions and Assistant Responses.}
Despite these developments, no existing system uses the functional distinction between user interactions and assistant responses as an explicit organizing principle. User inputs call for relational, graph-based representations that track changing preferences, facts, and context across sessions, whereas assistant responses are better decomposed into independently retrievable atomic units so that downstream models receive only the relevant excerpts. Scalaris closes this gap, unifying short- and long-term stores while applying a representation tailored to each source.

\begin{figure*}[t]
    \centering
    \includegraphics[width=0.9\textwidth]{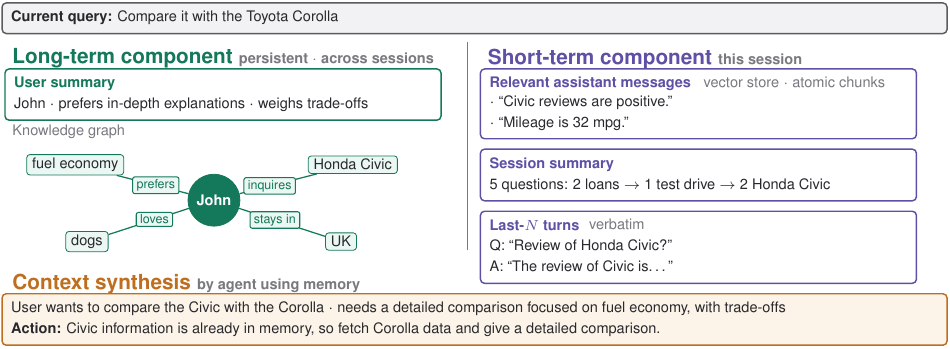}
    \caption{An illustrative instance of context retrieval in Scalaris. Long-term memory supplies the user's preference for fuel economy together with prior Honda Civic interactions, while short-term memory supplies the Relevant Assistant Messages, session summary, and recent turns (only the latest exchange is shown, for visual clarity). Together these resolve the referent, let the agent determine that only Corolla information must be retrieved, and enable personalization by focusing on fuel economy and including trade-offs.}
    \label{fig:hero}
\end{figure*}

\section{Methodology}
\label{sec:methodology}

\subsection{System Architecture}

Scalaris is a unified long-term and short-term memory architecture designed for conversational AI assistants. As illustrated in Figure~\ref{fig:hero}, the architecture is partitioned into a long-term component tailored for cross-session recall and a short-term component dedicated to intra-session continuity.

The \textbf{long-term component} is built on Graphiti~\citep{rasmussen2025zeptemporalknowledgegraph,graphiti}, an open-source temporal knowledge graph framework, which we substantially modify for use as a memory layer. It consists of a user summary and Relevant user interactions. The user summary maintains an evolving, holistic profile of the user's attributes, behavioral patterns, and focus areas. It is updated incrementally after each interaction and periodically rebuilt from historical logs to prevent semantic drift. Relevant user interactions extract three elements from user messages: \textit{Relevant Facts}, which capture temporal semantic data; \textit{Relevant Entities}, which represent previously discussed people, organizations, or concepts alongside their context; and \textit{Relevant Past Messages}, which preserve the verbatim user interactions.

\begin{figure*}[t]
    \centering
    \includegraphics[width=0.9\textwidth]{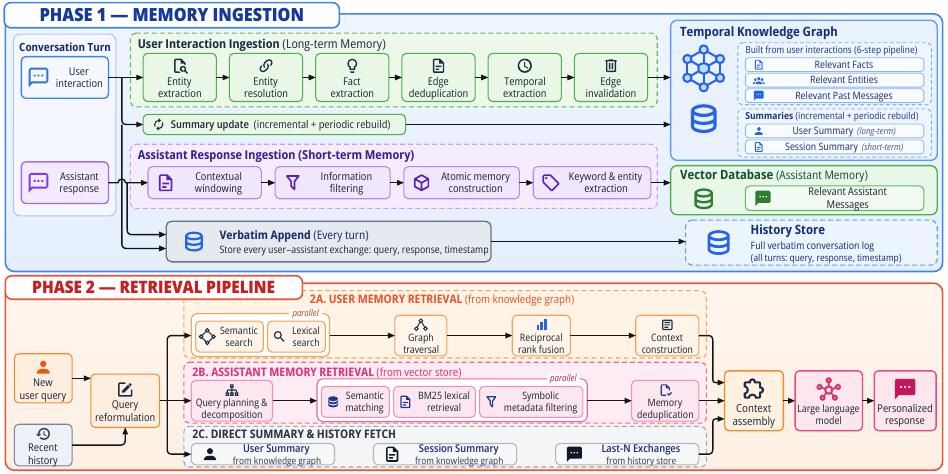}
     \caption{Flow diagram of the ingestion and retrieval pipeline of Scalaris. The ingestion phase organizes user interactions as facts, entities, and summaries in the graph store, assistant responses as atomic units in the vector store, and stores each incoming user query and AI response verbatim in the history store. The retrieval pipeline begins by reformulating the incoming query, which is then used to retrieve information from the graph and vector stores. Summaries are fetched directly from the graph store, and the $N$ most recent exchanges from the history store. All retrieved components are combined during context assembly and passed to the answering system.}
     \vspace{-1.0em}
    \label{fig:pipeline}
\end{figure*}

The \textbf{short-term component} consists of a session summary, the Last-$N$ Exchanges, and Relevant Assistant Messages. The session summary compactly tracks the ongoing dialogue state, while the Last-$N$ Exchanges provide the verbatim recent context required for immediate coreference resolution and intent tracking. For Relevant Assistant Messages, we leverage a vector database to store assistant responses as independently retrievable atomic units. The assistant's outputs are fragmented into self-contained units, ensuring only informative segments are retrieved rather than full, verbose responses. We intentionally keep Relevant Assistant Messages in short-term memory for two reasons. First, assistant responses are often not useful across sessions, and enabling cross-session retrieval increases retrieved context with limited benefit. Second, in domains with rapidly changing information like finance, retrieving stale assistant responses increases the likelihood of hallucinations. This behavior is nonetheless configurable for applications that require assistant responses to persist across sessions. Figure~\ref{fig:pipeline} details the complete ingestion and retrieval pipeline.

\subsection{Ingestion}

\paragraph{Relevant user interactions.}
Each incoming user message is ingested as an episode from which entities, facts, and temporal information are extracted and linked to the memory graph. The memory uses a bi-temporal model that tracks both a fact's validity interval and when it was learned or recorded, preserving historical changes. Each episode is processed as follows:

\begin{enumerate}[label=(\alph*)]
    \item \textbf{Entity Extraction.} Entities are extracted from each episode, and a concise summary of up to 250 characters is maintained for each entity to capture its conversational context.
    \item \textbf{Entity Resolution.} Extracted entities are matched against existing entities using a resolution pipeline that combines exact matching, lexical similarity, and embedding-based similarity. If a match is identified, the entity is linked to its existing graph representation; otherwise, a new representation is created.
    \item \textbf{Fact Extraction.} Atomic facts are extracted from the episode and represented as relationships between resolved entities, capturing the specific information conveyed by the user and its associated temporal context.
    \item \textbf{Edge De-duplication.} Newly extracted relationships are compared against existing edges to identify duplicate information, preventing repeated interactions from unnecessarily creating redundant relationships in the graph.
    \item \textbf{Temporal Extraction.} Temporal expressions associated with extracted facts are identified and normalized to determine the validity interval of each relationship, including explicit dates as well as relative expressions found in the conversation context.
    \item \textbf{Edge Invalidation.} When a newly extracted fact contradicts an existing relationship, the validity of the previous edge is terminated at the point when the new fact becomes valid, rather than deleting the historical edge, preserving the evolution of the user's information over time.
\end{enumerate}

For each extracted fact, the original user message is retained alongside its corresponding memory representation, preserving the source utterance for subsequent retrieval.

\paragraph{User and Session Summaries.}
User and session summaries are maintained in the same graph-based memory layer and follow a common update strategy. Both summaries are updated incrementally as new interactions are observed and periodically reconstructed from the underlying conversation history, with interactions provided in temporal order during reconstruction. Let $\sigma_t$ denote the summary at time $t$, $x_t$ the newly observed interaction at time $t$, and $H_t$ the underlying user interaction history available at the reconstruction point:
\begin{equation}
\small
\sigma_t =
\begin{cases}
\operatorname{Rebuild}(H_t), & \text{at reconstruction intervals}, \\
\operatorname{Update}(\sigma_{t-1}, x_t), & \text{otherwise}.
\end{cases}
\end{equation}

\paragraph{Relevant Assistant Messages.}
Assistant responses are fragmented into atomic memory units to retain only informative content from each response and enable retrieval of only the relevant portions when needed. This is achieved through the following steps:

\begin{enumerate}[label=(\alph*)]
    \item \textbf{Contextual Windowing.} The conversation is divided into fixed-size windows of dialogue, with each window providing the surrounding context required to interpret the assistant response being processed.
    \item \textbf{Information Filtering.} Each dialogue window is evaluated for the information content of its assistant response, and windows containing only greetings, acknowledgements, or other low-value interactions are discarded to avoid storing uninformative content.
    \item \textbf{Atomic Memory Construction.} Informative assistant responses are decomposed into self-contained atomic memory units with temporal awareness. During construction, references to preceding dialogue are resolved, and related information is combined when necessary to form a complete representation.
    \item \textbf{Keyword and Entity Extraction.} Relevant keywords and entities are extracted and stored alongside each memory unit as metadata.
\end{enumerate}

\paragraph{Last-\emph{N} Exchanges.}
All user-assistant messages are stored verbatim in a history store, preserving their exact wording and ordering.

\subsection{Retrieval}

\paragraph{Query Reformulation.}
Queries in ongoing conversations often depend on preceding dialogue through pronouns, ellipsis, or implicit references, making them incomplete retrieval keys in isolation. Each incoming query is therefore reformulated using recent conversation history before retrieval. Given the current query $q_t$ and recent conversation history $H_t^{\mathrm{recent}}$, a reformulation function produces a self-contained query
\begin{equation}
q_t^* = R(q_t, H_t^{\mathrm{recent}}).
\end{equation}
The reformulation resolves references, restores omitted subjects, and makes implicit continuations explicit. If $q_t$ is already self-contained, it is retained unchanged, such that $q_t^* = q_t$. The resulting query $q_t^*$ is used for all subsequent retrieval operations.

\paragraph{Relevant user interactions.}
Relevant user interactions are retrieved through a three-stage search, reranking, and context-construction pipeline.

\begin{enumerate}[label=(\alph*)]
    \item \textbf{Search.} The reformulated query is used to retrieve candidate facts and entities through semantic, lexical, and graph-traversal search. Semantic search uses dense embedding similarity, computed with Cohere's \texttt{embed-english-v3.0} model~\citep{cohere-embed-v3} to identify facts and entities that are semantically related to the query. Lexical search uses BM25~\citep{robertson2009probabilistic} over fact text and entity names to recover exact terms, names, and rare expressions. Graph traversal complements these searches by starting from seed entities and exploring a bounded number of graph hops, retrieving related facts that may not be lexically or semantically similar to the query.

    \item \textbf{Reranking.} Candidate results are reranked using Reciprocal Rank Fusion (RRF)~\citep{cormack2009reciprocal}, which combines independently ranked lists without requiring their scores to be directly comparable. For a candidate $d$, the fused scores for facts and entities are computed as:
    \begin{equation}
    R_{\mathrm{fact}}(d) = \sum_{r \in \{\mathrm{sem, lex, graph}\}} \frac{1}{k + \mathrm{rank}_r(d)},
    \end{equation}
    \begin{equation}
    R_{\mathrm{entity}}(d) = \sum_{r \in \{\mathrm{sem, lex}\}} \frac{1}{k + \mathrm{rank}_r(d)},
    \end{equation}
    where $d$ denotes a candidate result, $r$ denotes a retrieval ranking, $\mathrm{rank}_r(d)$ is the rank of $d$ in ranking $r$, and $k$ is the RRF smoothing constant, which we set to the default value of $k=60$. This stage improves precision by retaining the highest-ranked candidates across complementary retrieval mechanisms.

    \item \textbf{Context Construction.} The reranked results are converted into the context supplied to the answering system. Each relevant fact is represented with its descriptive text, while each relevant entity is represented by its name and concise summary. Relevant Past Messages can also be retrieved from the metadata of relevant edges to provide the assistant with the exact user utterances from which the corresponding knowledge was derived. 
\end{enumerate}

\paragraph{Relevant Assistant Messages.}

\begin{enumerate}[label=(\alph*)]
    \item \textbf{Query Planning and Decomposition.} Before retrieval begins, the reformulated query is passed to a single LLM call that performs all query-planning operations required by the subsequent retrieval stages. The LLM analyzes the information need and simultaneously generates one or more self-contained query phrasings for semantic retrieval, extracts keywords for lexical retrieval, and identifies attributes—including persons, locations, entities, and temporal expressions—for symbolic retrieval. Relative temporal expressions are resolved into absolute time ranges using the current date and time as a reference.

    \item \textbf{Hybrid Retrieval.} The three retrieval views execute concurrently, each using its corresponding query representation. Semantic retrieval ranks memory units by embedding similarity, lexical retrieval uses BM25 to identify exact or closely matching terms, and symbolic retrieval filters memory units using stored attributes such as persons, locations, entities, and temporal ranges:
    \begin{equation}
    \small
    \begin{aligned}
    R_{\mathrm{sem}}
    &= \operatorname{TopK}_{n_{\mathrm{sem}}}
       \Bigl\{\, m_i \,\Bigm| \\
    &\qquad \cos(E(q_{\mathrm{sem}}), E(m_i))
       \geq \tau_{\mathrm{sem}}
       \Bigr\}, \\
    R_{\mathrm{lex}}
    &= \operatorname{TopK}_{n_{\mathrm{lex}}}
       \{ m_i \mid \operatorname{BM25}(q_{\mathrm{lex}}, m_i)
       \geq \tau_{\mathrm{lex}} \}, \\
    R_{\mathrm{sym}}
    &= \operatorname{TopK}_{n_{\mathrm{sym}}}
       \{ m_i \mid \operatorname{Meta}(m_i)
       \models q_{\mathrm{sym}} \}.
    \end{aligned}
    \end{equation}
    Here, $m_i$ denotes a candidate memory unit; $q_{\mathrm{sem}}$, $q_{\mathrm{lex}}$, and $q_{\mathrm{sym}}$ denote the query representations generated for semantic, lexical, and symbolic retrieval, respectively; $E(\cdot)$ denotes the embedding function; $n_{\mathrm{sem}}$, $n_{\mathrm{lex}}$, and $n_{\mathrm{sym}}$ specify the number of candidates retained from each retrieval view; and $\tau_{\mathrm{sem}}$ and $\tau_{\mathrm{lex}}$ are the minimum similarity and BM25 score thresholds, respectively. $\operatorname{Meta}(m_i) \models q_{\mathrm{sym}}$ indicates that the stored metadata of $m_i$ satisfies the constraints specified by $q_{\mathrm{sym}}$.

    Duplicate memory units returned by multiple views are removed using their unique identifiers, producing a single consolidated set of Relevant Assistant Messages.
\end{enumerate}

\paragraph{Summaries and Last-\emph{N} Exchanges.}
The user and session summaries are directly fetched from their corresponding graph nodes, while the most recent $N$ user-assistant exchanges are fetched from the history store verbatim.

\paragraph{Context Assembly.}
Finally, the outputs of all memory components are concatenated and passed to the answering system.

\section{Experiments and Results}

\subsection{Experimental Setup}

\subsubsection{Datasets}
We evaluate Scalaris on two long-term conversational memory benchmarks: \textbf{LongMemEval-S}~\citep{wu2025longmemevalbenchmarkingchatassistants}, comprising 500 questions across six evaluation categories with approximately 115K input tokens per question, and \textbf{LoCoMo}~\citep{maharana2024evaluatinglongtermconversationalmemory}, comprising 1,986 questions across 10 multi-session conversations and five categories, with approximately 9K tokens per conversation. For LoCoMo, we follow the category labels and question distribution provided by the Hugging Face version of the dataset.

\subsubsection{Evaluation Metrics}
To assess both the effectiveness and practical viability of the memory systems, we measure \textbf{accuracy} (\%), the primary metric for retrieval and reasoning success across all benchmark categories; \textbf{retrieval efficiency}, the average number of context tokens retrieved per query; and \textbf{latency}, the average retrieval time per query in seconds.

\subsubsection{Implementation Details}
For the end-to-end memory system, we consider Qwen3-32B~\citep{yang2025qwen3technicalreport} and OSS-120B~\citep{openai2025gptoss120bgptoss20bmodel} as the underlying models. Following benchmark protocols, GPT-4o~\citep{openai2024gpt4osystemcard} is used as the answer generator and evaluator. 

Because these benchmarks primarily evaluate long-term memory recall and reasoning rather than conversational continuity, we exclude the session summary and Last-$N$ Exchanges during evaluation. We configure Relevant Assistant Messages for cross-session retrieval to align with the benchmarks' requirements. All other components, including the proposed ingestion and retrieval pipelines, remain unchanged from the core architecture. The recommended hyperparameter ranges and sensitivity analysis are presented in the extended analysis in Appendix ~\ref{sec:hyperparameter_sensitivity}. Baseline results in Tables~1 and~2 are taken from \citet{latimer2025hindsight}.

\subsection{Results}

\subsubsection{Performance on LongMemEval-S}
Table~\ref{tab:longmemeval} compares Scalaris with existing memory systems across six categories of LongMemEval-S.

\begin{table}[!ht]
\centering
\scriptsize
\setlength{\tabcolsep}{2.5pt}
\renewcommand{\arraystretch}{1.15}

\resizebox{\columnwidth}{!}{%
\begin{tabular}{lccccccc}
\toprule
\rowcolor{headergray}
\textbf{Type} &
\makecell{\textbf{Zep}\\[-1pt]{\tiny GPT-4o}} &
\makecell{\textbf{Superm.}\\[-1pt]{\tiny GPT-5}} &
\makecell{\textbf{Superm.}\\[-1pt]{\tiny Gemini-3}} &
\makecell{\textbf{Hinds.}\\[-1pt]{\tiny OSS-120B}} &
\makecell{\textbf{Hinds.}\\[-1pt]{\tiny Gemini-3}} &
\makecell{\textbf{Scalaris}\\[-1pt]{\tiny Qwen3-32B}} &
\makecell{\textbf{Scalaris}\\[-1pt]{\tiny OSS-120B}} \\
\midrule

\rowcolor{rowgray}
SSU & 92.9 & 97.1 & \underline{98.6} & \textbf{100.0} & 97.1 & 97.14 & 95.71 \\

SSA & 80.4 & \textbf{100.0} & \underline{98.2} & \underline{98.2} & 96.4 & 83.93 & 85.71 \\

\rowcolor{rowgray}
SSP & 56.7 & 76.7 & 70.0 & \underline{86.7} & 80.0 & \textbf{93.33} & 80.0 \\

KU & 83.3 & 87.2 & 89.7 & \underline{92.3} & \textbf{94.9} & 89.74 & 89.74 \\

\rowcolor{rowgray}
TR & 62.4 & 81.2 & 82.0 & 85.7 & \underline{91.0} & \textbf{93.98} & 90.98 \\

MS & 57.9 & 75.2 & 76.7 & 81.2 & \underline{87.2} & \textbf{92.48} & \textbf{92.48} \\

\midrule
\rowcolor{overallgray}
\textbf{Overall} & 71.2 & 84.6 & 85.2 & 89.0 & \underline{91.4} & \textbf{92.2} & 90.6 \\

\bottomrule
\end{tabular}%
}

\caption{Performance comparison on LongMemEval-S (\% Accuracy).
SSU: single-session-user; SSA: single-session-assistant;
SSP: single-session-preference; KU: knowledge-update;
TR: temporal-reasoning; MS: multi-session. Superm.: Supermemory; Hinds.: Hindsight.
(\textbf{Bold}: best; \underline{underline}: second best)}
\label{tab:longmemeval}
\end{table}


Scalaris achieves the highest overall accuracy of 92.2\%, outperforming all compared systems while utilizing an average of only 4,116 context tokens per query and maintaining a 0.94 s retrieval latency. It achieves state-of-the-art performance in three critical categories: SSP (93.33\%), TR (93.98\%), and MS (92.48\%). Notably, this 92.2\% overall accuracy is achieved with Qwen3-32B, a substantially smaller model than those used by the competing baselines. When paired with OSS-120B, Scalaris still outperforms most baseline configurations with 90.6\% accuracy, including Hindsight with the same backbone (89.0\%), while using 3,866 context tokens per query and 1.04 s latency. Failure cases are analyzed in Appendix~\ref{sec:error_analysis}.

\subsubsection{Performance on LoCoMo}
As most baseline frameworks do not evaluate adversarial questions, Table~\ref{tab:locomo} reports the four standard LoCoMo categories for direct comparison, with adversarial results detailed in the text.

\begin{table}[!ht]
\centering
\scriptsize
\setlength{\tabcolsep}{2pt}
\renewcommand{\arraystretch}{1.1}

\begin{tabular*}{\columnwidth}{l@{\extracolsep{\fill}}cccccc}
\toprule
\rowcolor{headergray}
\textbf{Category} &
\textbf{Zep} &
\makecell{\textbf{Mem0-}\\[-1pt]\textbf{Graph}} &
\textbf{Mem0} &
\makecell{\textbf{Lang-}\\[-1pt]\textbf{Mem}} &
\makecell{\textbf{Hind-}\\[-1pt]\textbf{sight}\\[-1pt]{\tiny Gemini-3}} &
\makecell{\textbf{Scalaris}\\[-1pt]{\tiny Qwen3-32B}} \\
\midrule

\rowcolor{rowgray}
S-Hop & 74.11 & 65.71 & 67.13 & 62.23 & \underline{86.17} & \textbf{89.36} \\

M-Hop & 66.04 & 47.19 & 51.15 & 47.92 & \underline{70.83} & \textbf{90.34} \\

\rowcolor{rowgray}
O-Domain & 67.71 & 75.71 & 72.93 & 71.12 & \textbf{95.12} & \underline{93.10} \\

Temporal & \underline{79.79} & 58.13 & 55.51 & 23.43 & \textbf{83.80} & 76.04 \\

\midrule
\rowcolor{overallgray}
\textbf{Overall} & 75.14 & 68.44 & 66.88 & 58.10 & \underline{89.61} & \textbf{90.77} \\
\bottomrule
\end{tabular*}

\caption{Performance comparison on LoCoMo (\% Accuracy).
S-Hop: Single-Hop; M-Hop: Multi-Hop; O-Domain: Open Domain.
}
\label{tab:locomo}
\end{table}

Scalaris yields the highest overall accuracy of 90.77\% among the compared systems and dominates the S-Hop (89.36\%) and M-Hop (90.34\%) categories. When the adversarial subset is included in the evaluation, Scalaris's overall accuracy rises to 92.55\%, driven by a 98.65\% accuracy on adversarial questions. On this benchmark, Scalaris uses an average of 4,145 context tokens per query with a retrieval latency of 0.98 s.

\subsubsection{Accuracy--Efficiency Trade-off}
Accuracy alone does not fully characterize the practical effectiveness of a memory system, as retrieved context length directly dictates enterprise inference costs. We evaluate this trade-off by comparing Scalaris with Hindsight (Table~\ref{tab:tokens-accuracy}). Because Hindsight publicly reports detailed retrieval metrics~\citep{hindsight-benchmarks}, the figures in Table~\ref{tab:tokens-accuracy} utilize these publicly reported values for a strictly fair comparison (and thus differ slightly from baseline figures in Tables~\ref{tab:longmemeval} and~\ref{tab:locomo}, which were taken from \citet{latimer2025hindsight}).

\begin{table}[!ht]
\centering
\small
\setlength{\tabcolsep}{3pt}
\begin{tabular}{lcccc}
\toprule
\rowcolor{headergray}
\textbf{Method} & \textbf{LME Acc.} & \textbf{LME} & \textbf{LoCoMo Acc.} & \textbf{LoCoMo} \\
\rowcolor{headergray}
& \textbf{(\%)} & \textbf{Tokens} & \textbf{(\%)} & \textbf{Tokens} \\
\midrule
\rowcolor{rowgray}
Hindsight & 94.6 & 43,625 & 92.0 & 36,235 \\
\rowcolor{overallgray}
\textbf{Scalaris} & \textbf{92.2} & \textbf{4,116} & \textbf{90.77} & \textbf{4,145} \\
\bottomrule
\end{tabular}
\caption{Tokens vs. accuracy (LME = LongMemEval).}
\label{tab:tokens-accuracy}
\end{table}

Scalaris achieves comparable accuracy to Hindsight while utilizing approximately 10$\times$ fewer retrieved context tokens on both benchmarks. Neither baseline distinguishes user interactions from assistant responses, and each sits at one end of the trade-off: Zep organizes all conversational content in a single knowledge graph, retrieving exceptionally lightweight contexts (averaging 1.6K tokens) at the cost of severe accuracy degradation (71.2\% on LongMemEval)~\citep{rasmussen2025zeptemporalknowledgegraph}, whereas Hindsight stores all content in a single PostgreSQL-backed memory bank~\citep{latimer2025hindsight}, securing high accuracy but demanding massive context windows. By representing each source according to its role, Scalaris attains comparable accuracy at a fraction of the context, validating our design choice to structurally differentiate user interactions from assistant responses.

\subsection{Ablation Study}
\label{sec:ablation}

We conduct an ablation study to isolate the impact of our dual-representation memory strategy. All experiments use Qwen3-32B as the underlying model, with retrieval hyperparameters held constant across configurations. We compare four configurations by alternating the memory structures used for user interactions and assistant responses. The \textit{Graph} configuration utilizes the ingestion and retrieval pipeline designed for Relevant user interactions, while the \textit{VectorDB} configuration follows the pipeline for Relevant Assistant Messages. Table~\ref{tab:ablation} reports the results across both benchmarks.
\begin{table}[!ht]
\centering
\scriptsize
\setlength{\tabcolsep}{1.5pt}
\renewcommand{\arraystretch}{1.15}

\begin{tabular*}{\columnwidth}{l@{\extracolsep{\fill}}lcccc}
\toprule
\rowcolor{headergray}
\multicolumn{2}{c}{\textbf{Memory organization}} &
\multicolumn{2}{c}{\textbf{LongMemEval-S}} &
\multicolumn{2}{c}{\textbf{LoCoMo}} \\
\cmidrule(lr){1-2}\cmidrule(lr){3-4}\cmidrule(lr){5-6}

\rowcolor{headergray}
\makecell[l]{\textbf{User}\\\textbf{interactions}} &
\makecell[l]{\textbf{Assistant}\\\textbf{responses}} &
\textbf{Acc.} & \makecell{\textbf{Avg.}\\\textbf{tokens}} &
\textbf{Acc.} & \makecell{\textbf{Avg.}\\\textbf{tokens}} \\
\midrule

\rowcolor{rowgray}
Graph & Graph & 76.2 & \underline{5,532} & 73.5 & \underline{5,729} \\

VectorDB & Graph & 79.3 & 6,267 & 77.2 & 6,183 \\

\rowcolor{rowgray}
VectorDB & VectorDB & \underline{86.3} & 8,729 & \underline{88.5} & 8,325 \\

\rowcolor{overallgray}
\textbf{Graph} & \textbf{VectorDB} &
\textbf{92.2} & \textbf{4,116} &
\textbf{90.77} & \textbf{4,145} \\

\bottomrule
\end{tabular*}

\caption{Ablation study of memory organization on LongMemEval-S and LoCoMo
(\% Accuracy). The last row is the proposed Scalaris organization.}
\label{tab:ablation}
\end{table}

The results demonstrate that memory organization plays a critical role in retrieval effectiveness and efficiency. Utilizing a graph-based structure for assistant responses yields poor accuracy despite retrieving more context, suggesting that relational graphs are not well suited for assistant responses. Conversely, using a VectorDB for both information streams provides substantially better accuracy, indicating that flat storage serves as a strong baseline. However, this homogeneous approach more than doubles the required context window while still underperforming the proposed architecture. 

Ultimately, by organizing user interactions via a temporal graph and assistant responses as atomic units in a VectorDB, Scalaris achieves state-of-the-art performance while requiring substantially fewer context tokens. These findings validate our core hypothesis: user and assistant messages exhibit distinct characteristics and require tailored memory representations. Additional retrieval-component ablations and hyperparameter sensitivity analyses are reported in Appendices ~\ref{sec:retrieval-component-analysis} and  ~\ref{sec:hyperparameter_sensitivity}.

\subsection{Internal Evaluation}

To validate Scalaris in a live production environment, we integrated the architecture into our proprietary, closed-source  conversational AI assistants and evaluated its performance on a sample of 1,200 real user interactions. For each query, we generated two responses: one with memory (WM) utilizing all Scalaris components, and one without memory (WOM) relying solely on the Last-$N$ Exchanges.

Using a blind evaluation protocol, GPT-4o scored each response on a 0--5 scale considering five dimensions: answer quality, personalization, clarity, actionability, and context consistency. Memory-enabled responses were preferred in 91.9\% of cases, achieving a mean score of 4.33 compared to 3.18 for the WOM baseline (+36.2\%). 
For the evaluation prompt and extended analysis, refer to the Appendices ~\ref{sec:Internal Evaluation Analysis} and ~\ref{sec: Internal Evaluation Prompt}.

Notably, the performance gap widened significantly as dialogue length increased. For queries in the latter half of extended sessions (defined as sessions exceeding 20 turns, representing 433 queries in our dataset), WM responses averaged 4.62, while WOM responses dropped sharply to 2.40. This confirms Scalaris's ability to maintain conversational coherence in long-horizon interactions.

\section{Conclusion}
We present Scalaris, a unified long-term and short-term memory architecture designed to address the diverse memory requirements of conversational AI systems. Scalaris enables agents to personalize responses based on users' preferences and interests, recall relevant information with temporal awareness, and maintain coherence across long and complex conversations. By appropriately organizing user interactions and assistant responses according to their characteristics, Scalaris achieves state-of-the-art performance on both LongMemEval-S and LoCoMo. Notably, these results are achieved with a relatively small Qwen3-32B model while requiring approximately 10$\times$ fewer retrieved context tokens than Hindsight with comparable accuracy. Internal evaluation further demonstrates significant improvements in response quality when Scalaris is integrated with conversational AI assistants, highlighting its potential as a practical and scalable framework for enterprise conversational AI systems.

\section{Limitations}

The current Scalaris architecture is restricted to declarative and episodic memory—such as user preferences, explicit facts, entities, and past interactions—and lacks dedicated mechanisms for modeling procedural knowledge like learned skills, workflows, or reusable action sequences. Furthermore, Scalaris is designed exclusively for text-based interfaces and has not been evaluated in voice-driven environments, which require strict real-time latency, continuous streaming, and low-overhead retrieval. \textbf{Future Work:} To address these constraints, future research will extend the framework to incorporate procedural memory capabilities while optimizing retrieval pipelines and streaming efficiency to support ultra-low-latency voice interactions.

\section{Ethical Considerations}

While Scalaris enhances personalization and continuity in conversational AI, persistent memory architectures introduce potential privacy and security risks. Long-term memory stores sensitive user information across sessions, making data protection critical. To mitigate these risks, production deployments must incorporate robust access controls, data anonymization, and user-driven mechanisms for inspecting, editing, or deleting stored memory entries (aligning with right-to-be-forgotten regulations). Furthermore, because long-term memory retains user preferences and traits, systems must safeguard against amplifying user biases or unintentionally storing personally identifiable information (PII).

\paragraph{Artifacts and Licenses.} Scalaris builds on Graphiti~\citep{rasmussen2025zeptemporalknowledgegraph,graphiti}, released under the Apache-2.0 license. We evaluate on LongMemEval~\citep{wu2025longmemevalbenchmarkingchatassistants} (MIT license) and LoCoMo~\citep{maharana2024evaluatinglongtermconversationalmemory} (CC BY-NC 4.0), both used solely for non-commercial research evaluation in accordance with their licenses. We verified that these licenses permit our use and the planned release of our code, which will retain the original license and copyright notices; benchmark data will not be redistributed. No internal user data will be released.

\section{Use of AI Assistants}
AI-based writing tools were used exclusively for grammar correction, proofreading, and language polishing of author-drafted text. All core concepts, methodology, experiments, analyses, and final phrasing remain the original work of the authors, who take full responsibility for the content of this manuscript.

\bibliography{custom}

\appendix
\newpage
\clearpage
\section{Extended Analysis}
\label{sec:extended_analysis}

To rigorously evaluate the individual contributions of Scalaris's modules and assess its robustness to hyperparameter variations, we present an extended analysis encompassing fine-grained ablations, sensitivity experiments, and a qualitative error breakdown.

\subsection{Retrieval Component Analysis}
\label{sec:retrieval-component-analysis}

To better understand the contribution of individual retrieval mechanisms, we evaluate the retrieval pipelines of Relevant Assistant Messages and Relevant User Interactions independently on the corresponding LongMemEval categories. Specifically, the assistant-response retrieval pipeline is evaluated only on the single-session-assistant category, while the user-interaction retrieval pipeline is evaluated only on the single-session-user category. These subsets isolate questions whose answers primarily depend on assistant responses or user interactions, respectively, allowing each retrieval component to be analyzed without interference from the other memory stream. Also, these questions require information distributed across multiple human-AI exchanges within one session, where omitting even a single relevant memory unit can lead to an incomplete or incorrect response. During each experiment, only the retrieval mechanism under evaluation is enabled while all other retrieval components within the same pipeline are disabled. Since graph traversal (BFS) requires an initial set of retrieved entities, it is evaluated only in combination with semantic retrieval or BM25 rather than as an independent retrieval mechanism. For reference, the complete retrieval pipeline achieves 83.93\% accuracy on single-session-assistant and 97.14\% on single-session-user.

The single-session-assistant (SSA) gap relative to some baselines in Table~\ref{tab:longmemeval} reflects a deliberate design trade-off. The SSA subset contains 56 questions, of which Scalaris answers 47 correctly (83.93\%). As noted above, these questions require combining details from several assistant responses within a session, so an answer is correct only if every relevant response is retrieved; missing even one yields an incomplete answer. This setting naturally favors systems that retrieve large portions of the conversation: Hindsight, for instance, retrieves roughly 10$\times$ more context than Scalaris (Table~\ref{tab:tokens-accuracy}). Scalaris instead retrieves compact atomic units under a fixed retrieval budget, and among low-context systems it remains stronger than Zep (83.93\% vs.\ 80.4\%).
\begin{table}[!ht]
\centering
\small
\setlength{\tabcolsep}{4pt}
\begin{tabular}{lccccc}
\toprule
\textbf{Retrieval Method} & \textbf{k=5} & \textbf{k=10} & \textbf{k=15} & \textbf{k=20} & \textbf{k=25} \\
\midrule
Semantic Search & 52.1 & 62.3 & \textbf{69.2} & 69.7 & 69.9 \\
Lexical Search  & 44.1 & 49.2 & \textbf{54.7} & 55.2 & 55.3 \\
Symbolic Search & 36.7 & \textbf{37.7} & 37.8 & 37.8 & 37.8 \\
\bottomrule
\end{tabular}
\caption{Retrieval analysis for Relevant Assistant Messages (Single-Session-Assistant). The optimal retrieval budget for each method (the elbow point beyond which gains become negligible) is shown in \textbf{bold}.}
\label{tab:retrieval-assistant}
\end{table}

As shown in Table~\ref{tab:retrieval-assistant}, semantic retrieval substantially outperforms lexical retrieval by capturing paraphrases and semantically related assistant responses that do not necessarily share surface-level vocabulary with the query. Lexical retrieval nevertheless remains valuable for recalling exact phrases, named entities, and numerical information, but its reliance on token overlap limits overall recall. Symbolic retrieval alone performs considerably worse because metadata filtering narrows the search space but cannot identify semantically related assistant responses, making it insufficient as a standalone retrieval strategy. Increasing the retrieval budget consistently improves performance for semantic and lexical retrieval until the optimal value, after which the gains become negligible despite retrieving additional memory units. Symbolic retrieval saturates even earlier because the number of metadata-matching candidates is inherently limited. These observations suggest that semantic, lexical, and symbolic retrieval capture complementary characteristics of assistant responses, motivating their integration in the final retrieval pipeline while avoiding unnecessary retrieval beyond the optimal budget.

\begin{table}[!ht]
\centering
\small
\setlength{\tabcolsep}{4pt}
\begin{tabular}{lccccc}
\toprule
\textbf{Retrieval Method} & \textbf{k=5} & \textbf{k=10} & \textbf{k=15} & \textbf{k=20} & \textbf{k=25} \\
\midrule
Semantic Search & 63.8 & 71.9 & 77.2 & \textbf{79.6} & 79.8 \\
BM25            & 59.4 & 66.8 & \textbf{72.5} & 72.9 & 73.0 \\
Semantic + BFS  & 74.6 & 80.9 & 84.2 & \textbf{85.8} & 85.9 \\
BM25 + BFS      & 67.5 & 74.8 & \textbf{79.8} & 80.3 & 80.4 \\
\bottomrule
\end{tabular}
\caption{Retrieval analysis for Relevant User Interactions (Single-Session-User). The optimal retrieval budget for each method (the elbow point beyond which gains become negligible) is shown in \textbf{bold}.}
\label{tab:retrieval-user}
\end{table}

As reported in Table~\ref{tab:retrieval-user}, semantic retrieval consistently outperforms BM25 because user queries often refer to previously discussed facts using different wording, allowing dense retrieval to identify semantically related interactions that lexical matching fails to capture. However, BM25 remains valuable for queries involving exact entity names, identifiers, or uncommon user-specific terminology, where lexical overlap provides a stronger retrieval signal than semantic similarity. Incorporating graph traversal improves both retrieval strategies by following the relationships of the initially retrieved entities and bringing together related facts that are distributed across multiple user interactions. The improvement is considerably larger when combined with semantic retrieval, as semantically retrieved entities provide more relevant starting points for graph traversal, whereas BM25 already retrieves highly specific matches that leave less additional information to be discovered through the graph. Similar to the assistant-response pipeline, increasing the retrieval budget consistently improves performance until the optimal value, after which the gains become negligible despite retrieving additional facts. Taken together, these observations suggest that the three retrieval mechanisms capture different aspects of user memory, motivating their integration in the final retrieval pipeline.

\subsection{Hyperparameter Sensitivity}
\label{sec:hyperparameter_sensitivity}

We further evaluate the sensitivity of Scalaris to its primary hyperparameters. During this analysis, the retrieval budget for each retrieval mechanism is fixed to its corresponding optimal value identified in Section~\ref{sec:retrieval-component-analysis}, allowing the effect of each hyperparameter to be analyzed independently. Table~\ref{tab:hyperparameter-sensitivity} summarizes the evaluated ranges and the corresponding recommended values. 

\begin{table}[!ht]
\centering
\small
\setlength{\tabcolsep}{4pt}
\renewcommand{\arraystretch}{1.3}

\resizebox{\columnwidth}{!}{%
\begin{tabular}{@{}>{\raggedright\arraybackslash}p{2.5cm}
                  >{\raggedright\arraybackslash}p{1.5cm}
                  >{\raggedright\arraybackslash}p{1.3cm}
                  >{\raggedright\arraybackslash}p{1.2cm}@{}}
\toprule
\textbf{Hyper-} & \textbf{Range} & \textbf{Optimal} & \textbf{Eval.} \\
\textbf{parameter} & \textbf{Evaluated} & \textbf{Range} & \textbf{Dataset} \\
\midrule
Semantic similarity threshold & 0.3--0.7 (step 0.1) & 0.4--0.6 & LME \\
BM25 threshold & 0.6--1.1 (step 0.1) & 0.9--1.0 & LME \\
Graph traversal depth & 1--5 hops & 3 hops & LME \\
Last-$N$ Exchanges & 1, 3, 5, 7, 10 & 5 & Internal Eval. \\
\bottomrule
\end{tabular}%
}

\caption{Evaluated ranges and recommended values for Scalaris's primary hyperparameters. LME: LongMemEval.}
\label{tab:hyperparameter-sensitivity}
\end{table}

Overall, Scalaris remains stable across a broad range of parameter values, indicating that its performance is not sensitive to minor variations in hyperparameter selection. We vary the semantic similarity threshold from 0.3 to 0.7 in increments of 0.1, observing consistently strong performance within the 0.4--0.6 range. Lower thresholds admit semantically weak matches, whereas higher thresholds filter relevant memories that differ in wording from the query. Similarly, we evaluate BM25 thresholds from 0.6 to 1.1, with performance remaining stable within the 0.9--1.0 range. Lower thresholds admit low-scoring lexical matches, whereas higher thresholds begin filtering lexical matches required for retrieval. Increasing the graph traversal depth improves retrieval up to three hops, after which additional hops primarily retrieve less relevant neighboring entities while increasing retrieval latency with negligible accuracy gains. Finally, evaluation on our internal benchmark shows that retaining the most recent five user-assistant exchanges provides sufficient conversational context for immediate coreference resolution, whereas larger context windows increase prompt length without improving response quality.

\subsection{Qualitative Error Analysis}
\label{sec:error_analysis}

To understand the limitations of Scalaris, we conducted a qualitative analysis of the remaining $\sim$8\% of failure cases in LongMemEval. The majority of errors stem from three distinct patterns. First, \textit{ambiguous temporal expressions} (e.g., ``a couple of quarters ago'') occasionally fail to resolve accurately when the surrounding context lacks explicit anchor dates. Second, \textit{distributed evidence retrieval} remains challenging when the information required to answer a query is scattered across multiple distant interactions, but the initial retrieval stage fails to surface all relevant memory units needed for complete reasoning. Finally, \textit{implicit fact negation} poses a challenge; when a user implicitly changes a previously stated preference without directly contradicting it, the temporal edge invalidation fails to trigger, causing the system to retrieve outdated context. Addressing these implicit reasoning bottlenecks remains a critical area for future work.

\subsection{Internal Evaluation: Detailed Analysis}
\label{sec:Internal Evaluation Analysis}
To complement the quantitative results presented in Section~4.4, we performed a qualitative analysis of  1{,}200 real user queries collected from our proprietary production finance assistants. Unlike benchmark datasets, these conversations represent genuine user interactions and therefore naturally contain incomplete follow-up questions, pronouns, ellipsis, and implicit references. Rather than restating the complete context in every turn, users progressively rely on the assistant's conversational memory, making later queries increasingly ambiguous when viewed independently. A representative conversation from our evaluation is shown below:

\begin{enumerate}
    \item Compare Tata Steel and Hindalco.
    \item Which one looks better for long-term investment?
    \item Compare it with Larsen \& Toubro.
    \item Which one is the weakest stock now?
    \item Replace the weakest stock.
    \item Which one looks safer?
    \item What are the downside risks?
    \item What are the upside drivers?
    \item What is the expected timeline?
    \item Summarize the key insights.
\end{enumerate}

Although the first few questions explicitly identify the companies under discussion, the subsequent queries depend almost entirely on the preceding conversation. References such as ``it'', ``the weakest stock'', ``the downside risks'', ``the upside drivers'', and ``the expected timeline'' cannot be interpreted correctly without resolving the conversational context established in earlier turns. Similar interaction patterns were consistently observed throughout the internal evaluation, reflecting how users naturally communicate with conversational assistants rather than how benchmark datasets are typically constructed.

To understand how existing memory frameworks handle such conversations, we additionally evaluated the same production queries using Zep and Hindsight. A common limitation observed across both systems was their inability to consistently resolve conversational coreferences and implicit follow-up queries before retrieval. Retrieval was frequently performed using the surface form of the latest user query, causing relevant memories associated with previously discussed entities to be missed. In contrast, Scalaris first reformulates each conversationally dependent query into a self-contained retrieval query by resolving pronouns, omitted entities, and implicit continuations before initiating memory retrieval.

Despite this shared limitation, the two frameworks exhibited different behaviors. Zep was able to provide limited personalization through its maintained user summary, but often failed to recall previously generated assistant responses. Consequently, the downstream assistant repeatedly invoked external tools to retrieve information that had already been fetched earlier in the conversation, increasing unnecessary computation and response latency. Hindsight recalled prior assistant information more reliably by retrieving substantially larger portions of the conversation, thereby reducing repeated tool invocations. However, this came at the cost of retrieving significantly more context tokens, frequently overwhelming the answering model with redundant information. Furthermore, because Hindsight does not explicitly maintain an evolving user profile, it was generally less effective at personalizing responses according to long-term user preferences.

These observations suggest that an effective conversational memory system must adapt to the way users naturally communicate, rather than expecting them to formulate fully self-contained queries at every turn. Real-world conversations are inherently incremental, with users relying on shared conversational context through pronouns, omitted entities, and implicit references. Furthermore, accurate recall alone is insufficient for a practical memory system. It must retrieve only the information necessary to answer the current query, avoiding redundant or irrelevant context. This is particularly important in enterprise deployments, where excessive retrieval increases latency, inference costs, and can overwhelm the downstream language model, increasing the likelihood of hallucinations.

\subsection{ Internal Evaluation Prompt}
\label{sec: Internal Evaluation Prompt}
The exact scoring prompt supplied to the GPT-4o judge for the blind pairwise evaluation described in Section~4.4 is shown in Figure~\ref{fig:eval-prompt}. Placeholders in braces are populated per query.

\begin{figure*}[t]
\centering
\begin{tcolorbox}[enhanced, width=\textwidth, colback=pback, colframe=pframe,
  boxrule=0.6pt, arc=3pt, left=10pt, right=10pt, top=8pt, bottom=8pt,
  title={Response Evaluation Prompt}, fonttitle=\sffamily\bfseries,
  colbacktitle=pframe, coltitle=white, fontupper=\small]

\psec{ROLE AND TASK}
You are an expert evaluator for AI assistant responses. Your task is to evaluate two responses to the same user question.

\medskip
\begin{pslotbox}{USER QUESTION}
\ph{question}
\end{pslotbox}

\psec{CONVERSATION CONTEXT}
Last 5 exchanges: \ph{last\_n\_formatted}\\[2pt]
Memory API Context: \ph{memory\_api\_context if memory\_api\_context else "No memory context"}

\medskip
\begin{tcbraster}[raster columns=2, raster equal height, raster column skip=8pt]
\begin{pslotbox}{RESPONSE A}
\ph{response\_a}
\end{pslotbox}
\begin{pslotbox}{RESPONSE B}
\ph{response\_b}
\end{pslotbox}
\end{tcbraster}

\psec{EVALUATION CRITERIA}
Consider the following dimensions when assigning an overall score:
\begin{itemize}[leftmargin=1.4em, itemsep=1pt, topsep=3pt, label=\textcolor{pslot}{\textbullet}]
  \item \textbf{Answer Quality:} Accuracy, completeness, and factual correctness.
  \item \textbf{Personalization:} Effective use of relevant memory context.
  \item \textbf{Clarity:} Readability, organization, and coherent presentation.
  \item \textbf{Actionability:} Practicality and usefulness of recommendations.
  \item \textbf{Context Consistency:} Alignment with conversation history and prior instructions.
\end{itemize}
Assign each response a single overall score from 0 to 5.

\psec{REQUIRED OUTPUT FORMAT (JSON ONLY)}
\begin{tcolorbox}[enhanced, colback=white, colframe=gray!30, boxrule=0.4pt,
  arc=2pt, left=6pt, top=3pt, bottom=3pt, fontupper=\small\ttfamily]
\{\\
\hspace*{1em}"score\_a": 2.2,\\
\hspace*{1em}"score\_b": 3.1,\\
\hspace*{1em}"reasoning": "...",\\
\hspace*{1em}"verdict": "response\_b\_better"\\
\}
\end{tcolorbox}
The verdict must be exactly one of \texttt{"response\_a\_better"}, \texttt{"response\_b\_better"}, or \texttt{"equal"}. Return only the JSON object, without additional commentary.

\end{tcolorbox}
\caption{Evaluation prompt used by the GPT-4o judge to compare two AI assistant responses. The prompt provides the original user question, recent conversational history, and available memory context, followed by both candidate responses. The judge assesses answer quality, personalization, clarity, actionability, and consistency with the conversation before assigning a score from 0 to 5 to each response and producing a comparative verdict in a strictly specified JSON format.}
\label{fig:eval-prompt}
\end{figure*}

\end{document}